\documentclass{article}
\usepackage{arxiv}
\usepackage[utf8]{inputenc} 
\usepackage[T1]{fontenc}    
\usepackage{hyperref}       
\usepackage{url}            
\usepackage{booktabs}       
\usepackage{amsfonts}       
\usepackage{nicefrac}       
\usepackage{microtype}      
\usepackage{lipsum}		
\usepackage{graphicx}
\usepackage{amsmath,amssymb,amsfonts}
\usepackage{algorithmic}
\usepackage{graphicx}
\usepackage{textcomp}
\usepackage{xcolor}
\usepackage{booktabs} 
\usepackage{graphicx}
\usepackage{epstopdf}

\usepackage{url}
\usepackage{algorithmic}
\usepackage{hyperref}
\usepackage{multirow}
\usepackage{caption}
\usepackage{amssymb}
\usepackage{amsfonts}
\usepackage{float}
\usepackage{bm}
\usepackage{dsfont}
\usepackage[T1]{fontenc}
\usepackage[left,modulo]{lineno}
\usepackage[ruled, vlined, linesnumbered, longend]{algorithm2e}
\usepackage{psfrag}
\usepackage{subfigure}
\usepackage{fancyhdr}
\usepackage{eucal}
\usepackage[english]{babel}

\usepackage{ifpdf}
\usepackage{setspace}
\usepackage{textcomp}
\usepackage{tikz}
\usepackage{tikz-inet}
\usetikzlibrary{calc,shapes.geometric}
\usetikzlibrary{shapes,snakes,arrows,automata} 
\usetikzlibrary{patterns}

\newif\ifnotes\notestrue
\def\boxnote#1#2{\ifnotes\fbox{\footnote{\ }}\ \footnotetext{ From #1:
#2}\fi}

\newcommand{\msb}[1]{{\color{blue}#1}}

\newcommand{\ben}{\begin{enumerate}}
\newcommand{\een}{\end{enumerate}}

\newcommand{\bc}{\begin{center}}
\newcommand{\ec}{\end{center}}

\newcommand{\bit}{\begin{itemize}}
\newcommand{\eit}{\end{itemize}}

\newcommand{\ds}{\displaystyle}
\newcommand{\beq}{\begin{equation}}
\newcommand{\eeq}{\end{equation}}

\newcommand{\wre}{\mathbf{W}^{\rm{h}}}

\newcommand{\wfb}{\mathbf{W}^{\rm{fb}}}

\newcommand{\vb}{\mathbf{b}}

\newcommand{\vy}{\mathbf{y}}

\newcommand{\y}{\mathbf{y}}

\newcommand{\Na}{n}

\newcommand{\Y}{\mathbf{Y}}

\title{Tracking changes using Kullback-Leibler divergence for the
continual learning
\thanks{This work was supported by the CEUS-UNISONO programme, which has received funding from the National Science Centre, Poland under grant agreement No. 2020/02/Y/ST6/00037, and the GACR-Czech Science Foundation project No. 21-33574K ``Lifelong Machine Learning on Data Streams''.
Authors are also thanksful to Pawe{\l}~Zyblewski for his helpful constructive feedbacks.
}
}

\author{\href{https://orcid.org/0000-0002-9172-0155}{\includegraphics[scale=0.06]{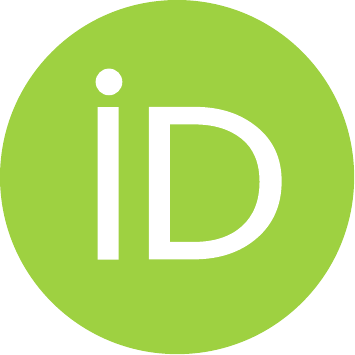}\hspace{1mm}Sebasti\'an Basterrech}\\
	Faculty of Electrical Engineering and Computer Science\\
	V\v{S}B-Technical University of Ostrava\\
	Ostrava, Czech Republic\\
	\texttt{Sebastian.Basterrech@vsb.cz} \\
	\And
	\href{https://orcid.org/0000-0003-0146-4205}{\includegraphics[scale=0.06]{orcid.pdf}\hspace{1mm}Micha\l~Wo\'zniak} \\
	Department of Systems and Computer Networks\\
	Wroclaw University of Science and Technology\\
Wroclaw, Poland \\
	\texttt{Michal.Wozniak@pwr.edu.pl}\\
}

\date{}
\renewcommand{\headeright}{$\;$}
\renewcommand{\undertitle}{(Accepted manuscript at SMC'2022)}
\renewcommand{\shorttitle}{Tracking changes using KL divergence for the continual learning}

\hypersetup{
pdftitle={Tracking changes using Kullback-Leibler divergence for the
continual learning},
pdfsubject={cs.NE,cs.AI,cs.LG},
pdfauthor={Sebasti\'an Basterrech, Micha\l~Wo\'zniak},
pdfkeywords={Drift detection,  Continual Learning, Relative Entropy, Data Stream Learning, Lifelong Learning},
}

\begin{document}
\newcommand{\op}{w^+}
\newcommand{\om}{w^-}
\newcommand{\Rho}{\mathrm{P}}
\newcommand{\wresp}{\mathbf{w}^+}
\newcommand{\wresn}{\mathbf{w}^-}
\renewcommand{\wre}{\mathbf{W}^{\rm{r}}}
\renewcommand{\wfb}{\mathbf{W}^{\rm{fb}}}
\newcommand{\ve}{\mathbf{v}}
\newcommand{\noise}{\bm{\varepsilon}}
\newcommand{\idct}{\phi}
\newcommand{\chr}{\kappa}
\newcommand{\phenotype}{\chi}
\newcommand{\mapping}{\Psi}
\renewcommand{\dim}{M}
\renewcommand{\Na}{K}
\newcommand{\Nx}{N}
\newcommand{\Ny}{L}
\newcommand{\State}{\mathbf{S}}
\newcommand{\pos}{\mathbf{p}}
\renewcommand{\vec}{\mathbf{v}}
\newcommand{\best}{\mathbf{b}}

%

%
\newif\ifnotes\notestrue
\def\boxnote#1#2{\ifnotes\fbox{\footnote{\ }}\ \footnotetext{ From #1: #2}\fi}

\def\bgr#1{\boxnote{Gerardo}{\color{red}#1}}
\def\hgr#1{}
\newcommand{\mgr}[1]{{\color{red}#1}}

\def\bsb#1{\boxnote{Sebasti\'an}{\bf\color{blue}#1}}
\renewcommand{\msb}[1]{{\bf\color{blue}#1}}

\newcommand{\todo}[1]{{\color{brown}\textsf{\bf [#1]}}}

\newcommand{\forceindent}{\leavevmode{\parindent=2em\indent}}

\newcommand{\ti}{\!\times\!}
\newcommand{\dimwa}{\Nx\ti\Na}
\newcommand{\dimwr}{\Nx\ti\Nx}
\newcommand{\dimwfb}{\Nx\ti\Ny}
\newcommand{\dimwo}{\Nx\ti(\Na\!+\!\Ny)}

\newcommand{\espacio}[1]{}

\newcommand{\vu}{\mathbf{u}}
\renewcommand{\vy}{\mathbf{y}}
\newcommand{\vz}{\mathbf{z}}
\newcommand{\U}{\mathbb{U}}
\renewcommand{\Y}{\mathbb{Y}}
\newcommand{\M}{\mathcal{M}}
\newcommand{\pred}{\hat{\mathbf{y}}}
\newcommand{\param}{\bm{\theta}}

\maketitle

\begin{abstract}
Recently, continual learning has received a lot of attention. One of the significant problems is the occurrence of \emph{concept drift}, which consists of changing probabilistic characteristics of the incoming data. In the case of the classification task, this phenomenon destabilizes the model's performance and negatively affects the achieved prediction quality. 
Most current methods apply statistical learning and similarity analysis over the raw data. However, similarity analysis in streaming data remains a complex problem due to time limitation, non-precise values, fast decision speed, scalability, etc. 
This article introduces a novel method for monitoring changes in the probabilistic distribution of multi-dimensional data streams. As a measure of the rapidity of changes, we analyze the popular Kullback-Leibler divergence.
During the experimental study, we show how to use this metric to predict the concept drift occurrence and understand its nature. 
The obtained results encourage further work on the proposed methods and its application in the real tasks where the prediction of the future appearance of concept drift plays a crucial role, such as predictive maintenance.
\end{abstract}

\keywords{Drift detection \and  Continual Learning \and Relative Entropy \and Data Stream Learning \and Lifelong Learning}

\section{Introduction}
%
One of the critical problems in the analysis of streaming data is the possibility of changing the probabilistic characteristics of the task during the prediction model run. This phenomenon, called \emph{concept drift}~\cite{Widmer:1996}, may deteriorate the classification quality of the using predictor, and usually, its occurrence is unpredictable. Its appearance is typical in many daily decision-making tasks, as fraudsters may change the content of the e-mail to get past spam filters.

We may propose several \emph{concept drift} taxonomies. The first one focuses on how the drift impacts the probability characteristics of the learning task. If it changes decision boundary shapes, i.e., the \emph{posterior} probabilities have been changed \cite{Schlimmer:1986} then we face with so-called \emph{real concept drift}. \emph{Virtual drift} does not change the decision boundary shape. However, it changes the unconditional probability density function \cite{Widmer:1993}. Nevertheless, Oliveira et al. \cite{Oliveira:2021} pointed out that while \emph{virtual concept drift} does not change the shape of decision boundaries, it can reduce the usefulness of the decision boundaries used by classifiers.

Another taxonomy considers the change speed. Mainly, we may distinguish sudden \emph{concept drift}, when a new concept abruptly replaces an old one; and incremental \emph{concept drift}, when we may observe a steady progression from an old concept toward a new one. We should also mention gradual \emph{concept drift}, when during the transition between an old and a new concept, the two concepts may occur with different intensities. An interesting phenomenon is periodic changes, referred to as recurring \emph{concept drift} \cite{Sobolewski:2013}, when previously occurring concepts may recur, which is very typical for seasonal phenomena (cyclic concept \cite{Hoens:2012}).

Currently, most approaches are reactive, i.e., they focus on the problem of concept drift detection or continuous model adaptation to emerging changes. These approaches include concept drift detectors, which act as triggers for significant shifts in the probability distribution, and methods that continuously try to adapt to changes in the probability distribution. The second group of the mentioned methods mainly use the classifier ensemble \cite{Krawczyk:2017}, where the model consistency with the current probability distribution is ensured either by changing the composition of the classifier ensemble or updating individual base classifiers.

This paper focuses on the \emph{concept drift} monitoring problem and proposes a novel drift detection method. The significant advantage of the proposed approach is that it can compare distributions of unknown types to detect a drift without any assumptions about distribution parameters or distribution types.


%
%
This work offers the following contributions:
\begin{itemize}
    \item Using Kullback-Leibler divergence for ongoing monitoring of changes in probability distributions. 
    \item Introducing a novel \emph{concept drift} detector based on the mentioned metric over the raw data, so-called KL-divergence-
based concept drift detector (KLD).
    \item Fast and robust decision rule with a control real-value parameter that can be dynamically adjusted for improving performance of the matching matrix.
    \item Initial experimental evaluation of the proposed method for non-stationary data streams.
\end{itemize}

The remainder of the paper is organized as follows. The next section presents a background over the main concepts of this article. Section~\ref{Methodology} fully describes our contribution. Experimental study is presented in Section~\ref{ExpStudy}. We close with a summary of our contributions and discussing directions for future work.

\section{Preliminaries}
The following section presents the key information about data stream, concept drift, and Kullback-Leibler divergence necessary to explain the proposed approach.

\subsection{Streaming data}
Continual learning of streaming data is usually  associated to problems, where datasets become available in chunks.  
A data stream often is defined as an ordered sequence of chunks  $\{S_1, S_2,\ldots, S_k, \ldots\}$, where the chunk $S_i$ consists in a finite data sequence $S_i=\{\vz_i^{(1)},\vz_i^{(2)}, \ldots, \vz_i^{(K)}\}$.
Since our goal is to analyze continual learning over a data stream in a supervised context, then we assume data points~$\vz_i^{(k)}$ defined as input-output pairs $(\vu_i^{(k)}, \vy_i^{(k)})\in \U\times\Y$. 
The input vectors are in a $p$-dimensional space $\U$, and the  outputs belongs to a $d$-dimensional space $\Y$.
In case of a classification problem, the space $\Y$ describes a small number of classes. Otherwise, it is a regression problem  where $\Y$ is typically a subset in a $d$-dimensional real space.

Tracking changes in a data stream often are made using drift detection methods. Most of the drift detection techniques use a base predictor to classify incoming instances.
In general, the methods work as follows. For each input instance the base predictor outputs a class label which is compared to the true class label.
Then, the accuracy is evaluated and used as tool for deciding whether a drift has occurred or not.
There are several metrics to evaluate the accuracy of such prediction.
The choice of the selected metric, conventionally called loss function, depends on both the domain of $\Y$ and the optimization method applied for adjusting the base predictor.
Some commonly used functions are $0/1$ loss function and  cross-entropy in the case of classification problems and quadratic errors in the case of regression problems~\cite{Shaker:2015}.

\subsection{Drift detection}
%
A concept drift detector is an algorithm that can inform on data distribution changes. To detect a real drift, the labeled data or a classifier's performance are usually required, but also some proposition employing unlabeled data only, usually based on statistical tests as Smirnov-Kolmogorov are also proposed \cite{Sobolewski:2013}. Nevertheless, we should be aware of its limitations. Such detectors are good at detecting virtual concept drift and are not guaranteed to detect real drift. However, changes in the unconditional probability density distributions are rarely not accompanied by changes in the conditional distributions. Moreover, as mentioned in \cite{Oliveira:2021}, virtual drift often affects the utility of the prediction models used.

Some drift detectors can also return a warning signal, i.e., the distribution changing, but the change is insignificant. A warning signal could trigger collecting new data for upcoming updates or rebuilding the current predictor. 

Detecting the drift should be done as quickly as possible to replace an outdated model, thus minimizing the restoration time. On the other hand, false positives are also unacceptable because they can cause the model to be corrected unnecessarily, resulting in model adjustment to unrepresentative samples and increased, unwarranted consumption of computational resources \cite{Gustafsson:2000}. 

Nevertheless, according to the authors' best knowledge, the recommended drift detector's testing method should focus on the predictive performance of the classifier that employs the examined detectors. It is recommended to use the following framework~\cite{Goncalves2014}. 
%

Let us shortly review the most important drift detection algorithms. \emph{\textsc{cusum} (Cumulative sum)}~\cite{Page:1954} is a simple sequential analysis technique based on the measurement of the mean value of the input data. If the mean is significantly bigger than zero drift is. 
\emph{Exponentially Weighted Moving Average (\textsc{ewma})} \cite{Ross:2012} combines current and historical observations that allow detecting changes in the mean value quickly by using aggregated charting statistics. A weight factor is introduced that promotes the most current observations. 
\emph{\textsc{ddm} (Drift Detection Method)} \cite{Gama:2004} incrementally estimates a classifier performance, which (assuming the convergence of the learning algorithm) has to diminish along with new instances from the given distribution being continuously presented to the learner \cite{Raudys:2014}. If the reverse behavior is observed, we may suspect a change of probability distributions. \emph{{\textsc{eddm}} (Early Drift Detection Methods)} \cite{baena2006early} extends the \textsc{ddm}. It proposed the heuristic window size selection procedure, used different performance metric to follow, and implemented new warning and drift levels .
Blanco et al. \cite{Blanco:2015} propose detectors that use the non-parametric estimation of classifier error employing Hoeffding's and McDiarmid's inequalities. \emph{\textsc{adwin}} \cite{Bifet:2006} employs an adapting sliding window, being highly suitable for handling sudden drifts. When no change is apparent, it automatically grows the window size and shrinks it when data stream distribution changes. It tests the hypothesis about the average equality in two subwindows obtained using Hoeffding bounds.
Nishida and Yamauchi \cite{Nishida:2007} developed  \emph{\textsc{stepd}} that employs statistics, equivalent to the chi-square test with Yates's continuity correction to infer whether the classification quality of the current observations had changed since the current model was used.  
We have also mention the compound detection models which employ ensemble approach, that could be found, e.g., in \cite{Maciel:2015}, \cite{du2014selective}, \cite{Lapinski:2018}.
Concept drift detection has also been developed over high-dimensional and sparse spaces (e.g. sparse time series)~\cite{Zhang2019,Shang2022}.

%
%
%
%

\subsection{Kullback-Leibler divergence}
\label{KLdivegenceSection}
A function for assessing the similarity between two distributions is the relative entropy, so-called, \emph{Kullback-Leibler (KL)} divergence.
%
%
The KL divergence between two probability mass functions (pmf) $p(x)$ and $q(x)$ over a set $\mathbb{X}$ is defined as follows~\cite{Dasu2006}:
\begin{equation}
\label{KL}
    {\rm{KL}}(p(x)\parallel q(x))=\ds{\sum_{x\in\mathbb{X}}p(x)\log\frac{p(x)}{q(x)}},
\end{equation}
where the $\log(\cdot)$ is logarithmic function with an arbitrary selected log base (most commonly used $2$ or $e$).

%
%
Expression~(\ref{KL}) is zero if and only if both distributions have identical quantities of information.
An interpretation coming from \textit{Information Theory} is that $KL(p(x)\parallel q(x))$ gives the information lost when $q(x)$ is used to approximate $p(x)$.
When both distributions are strictly positive, then KL divergence value is always a positive real number.
However, expression~(\ref{KL}) is problematic when either $p(x)$ or $q(x)$ has zero values.
The expression goes to an infinity value when $p(x)$ tends to zero, and it may takes undefined value when $q(x)=0$.
A common convention is to define $0 \log(0) = 0\log(0/0)=0$ and $x\log(x/0)=\infty$ if $x>0$~\cite{Csiszar1998}.
A common strategy for avoiding numerical problems consists in applying a smoothing approach of the empirical distribution, which is done by adding a stable fixed noise to each element with a probability mass equal to zero.
%
%
Another important property of the KL divergence is that is not symmetric. Neither satisfies triangle inequality for any pair of distributions $p(x)$ and $q(x)$~\cite{Cover2012}.
Therefore, it is not strictly a metric.
As a consequence, it can be inconvenient to be used in some applications~\cite{Dasu2006}.
%
%

%
In spite of the described drawbacks, the KL divergence has also several advantages that makes expression~(\ref{KL}) suitable for being used as a similarity measure between two distributions. 
Here, we further discuss some of those helpful properties~\cite{Cover2012,Dasu2006}.
\begin{itemize}
\item Relationship with likelihood ratio: 
KL divergence is often presented as the expected value of the log likelihood ratio,
\begin{equation}
\label{KLdef2}
    {\rm{KL}}(p(x)\parallel q(x))=
    \mathbb{E}_p\bigg(\ds{\log\frac{p(x)}{q(x)}}\bigg).
\end{equation}
This definition can be specially useful when we know the distribution from the data, and it is easy to compute the log likelihood ratio. 
\item Relationship with Maximum Likelihood Estimation (MLE): MLE can be viewed as minimizing the KL divergence between the empirical distribution data and the proposed model distribution.
\item Hypothesis tests: 
several hypothesis tests are equivalent to KL divergence. The statistic of t-test is the KL distance between two Gaussian distributions. The first term of  the Taylor expansion of KL divergences is equivalent to the statistic of chi-square test.
Furthermore, the Kulldorff spatial test used for clustering data has statistic that is the KL-distance between the empirical data distribution and the Poisson distribution.

\item Pinsker's inequality: total variation distance between two densities, so called statistical distance, it is computed as a function of KL divergence of those two densities.


\end{itemize}

%

\section{KL-divergence-based concept drift detector (KLD)}
\label{Methodology}
%
%

This section presents a pipeline for a KL-divergence-based concept drift detector (KLD). Firstly, we show how to determine the empirical distribution from streaming data, calculate KL-divergence between two data chunks, and detect the drift using a simple threshold scheme.

\subsection{Creation of empirical distribution from a data stream}
Each chunk defines an empirical distribution, and we compute the similarity between two adjacent chunks using KL divergence.
Following the previous notation, let's define the data presented in chunk $S_i$ as a sequence of input-output pairs $\mathbf{z}_i^{(k)}=(\mathbf{u}_i^{(k)},\mathbf{y}_i^{(k)})$, 
where $\mathbf{z}^{(1)}_{i}, \mathbf{z}^{(2)}_{i}, \ldots, \mathbf{z}^{(K)}_{i}$ is a random sample from a discrete distribution with an unknown probability mass function over a discrete set $\U\times\Y$.
%


We assume unidimensional output space.
For the probability mass function estimation, we divide the input space $\U$ in a multidimensional regular grid composed by $J$ bins. 
Let $\vb_i^{j}$ be the bin $j$ in the chunk $i$.
A bin can be formalized defining a parameterized range constraint, and each point $\vu\in\mathbb{U}$ could be assigned to a specific bin according to the parameterized range constraint.
%
%
%
%
In a chunk $S_i$ the conditional probability of having a class $l$ in bin $\vb_i^{j}$ is estimated by  
\begin{equation}
\label{conditionalProb}
\Pr\big(\y_i=l|\{\vu_i:\vu_i^{k}\in \vb_i^{j}, \forall k\}\big)=
\displaystyle{
\frac{\sum_{k=1}^{K} \mathds{1}_{\{\vu_i^{k} \in \vb_i^{j}\}\cap \{\vy_i^{k}=l\}}}{\sum_{k=1}^{K}
\mathds{1}_{\{\vu_i^{k}\in \vb_i^{j}\}}
}},
\end{equation}
where  $\mathds{1}_{\{A\}}$ is $1$ if predicate $A$ is true, 0 otherwise.
Then, by applying expression~(\ref{conditionalProb}) for each output class $l$ is computed the probability mass function $\Pr(\vy_i |\{\vu_i:\vu_i^{k}\in \vb_i^{j}, \forall k\}\big)$.

\subsection{Proposed KL-divergence-based similarity metric}
Hence, each bin has associated a probability mass function, then we may see changes through any two chunks $S_i$ and $S_t$ computing 
\begin{equation}
\begin{aligned}
\label{distanceBins}
d^{j}_{i,t}=KL\big(\Pr(\vy_i |\{\vu_i:\vu_i^{k}\in \vb_i^{j}, \forall k\}\big)
\parallel
\Pr(\vy_{t} |\{\vu_t:\vu_t^{k}\in \vb_t^{j}, \forall k\}\big).
\end{aligned}
\end{equation}
Finally, similarity between any two chunks~$S_i$ and $S_{t}$ is defined 
\begin{equation}
\label{distanceChunks}
D(S_i,S_{t})=\ds{\frac{1}{J}\sum_{j=1}^{J} d_{i,t}^{j}}.
\end{equation}
Note that, in expression~(\ref{distanceChunks}) each bin has the same relevance in the total sum.
We present a slight variation that considers the probability of sampling points in each bin.
Then, the proposed similarity measure for comparing two chunks is given by the weighted sum 
\begin{equation}
\label{weightedDistanceChunks}
D(S_i,S_{t})=\ds{\frac{1}{J}\sum_{j=1}^{J}\gamma_i^{j} d_{i,t}^{j}},
\end{equation}
where the weight $\gamma_i^{j}$ is the probability estimation of sampling in a specific region $\vb_i^{j}$ of the reference chunk $S_i$
$$
\gamma_i^{j}= \frac{1}{K}\displaystyle\sum_{k=1}^{K}\mathds{1}_{\{\vu_i^{k}\in \vb_i^{j}\}}.
$$

\subsection{Decision rule using a threshold value}

In this first work, we analyze similarity between consecutive chunks. 
Therefore, we apply expression~(\ref{weightedDistanceChunks}) over $\{S_1, S_2,\ldots, S_k, \ldots\}$, then we obtain a sequence of positive reals 
$\{D(S_1, S_2),D(S_2, S_3) \ldots, D(S_k,S_{k+1}) \ldots\}$.
Intuitively, when $D(S_k,S_{k+1})$ value is \textit{large} (much larger than previous ones), then the proposed metric has identified a concept drift.
On the other hand, low values of $D(S_k,S_{k+1})$ represents unchanged distribution of the data.
However, important changes between two consecutive chunks does not necessarily mean that has occurred a concept drift. For instance, it can be just due to the occurrence of outliers in one chunk and not in the next one. Even, it can also occurs because the data stream is provided in small buffers (chunk size too small).
As a consequence, we smooth out the sequence $\{D(S_1, S_2),D(S_2, S_3) \ldots, D(S_k,S_{k+1}) \ldots\}$ 
 for mitigating the impact of outliers in sequential data applying a  smoother~\cite{Buja1989}.
In the experimental part, we analyzed two classic operations: moving average (with a time windows of~$5$ chunks) and a Locally Weighted Scatterplot Smoothing~\cite{Buja1989}.  
By definition, the smoothing mapping provides a curve with continuous first  derivative (at least first derivative is continuous).
We created a really fast and simple decision rule based on a threshold over the first derivative of the smoothed curve.
Let's denote the points of the first derivative of the smoothed curve by $\{l_1,l_2,\ldots \}$. 
We identify a \textit{critical} point location (i.e., timestamps when a concept drift occurs) when a gradient point 
\begin{equation}
\label{DecisionRule}
l_i\notin[\bar{l}-\alpha\sigma(l),\bar{l}+\alpha\sigma(l)],    
\end{equation}
 where $\bar{l}$ denotes the mean of the sequence, $\sigma(l)$ is the standard deviation, and $\alpha$ is a real-value control parameter. %

 \subsection{Main parameters}
\begin{itemize}
\item \textbf{Control of KL-divergence numerical  stability.} In cases of having pmf values equal to zero, then is necessary to make a smooth correction (using an arbitrarily small $\varepsilon>0$ parameter) as described in section~\ref{KLdivegenceSection}. 
\item \textbf{Estimation of probability mass function.} The estimation is based on the partition of the input space. The proposed partition is done using a regular grid with $J$ bins. This parameter may impact on the probability mass estimation and in the KL-divergence numerical stability. 
If $J$ is large, then it may exist bins without any data instance. As a consequence, $J$ large can increase the instability of the KL divergence computation. 
In addition, it impacts in the computational cost of the pmf estimation. 
To estimate a pmf over a regular grid with many bins is much more costly than to split the input space in few sub-regions. After carrying out multiple runs, on our experiments we fixed the number of bins to $5\times p$, where $p$ is the dimension of $\U$.
\item \textbf{Control of the smoothing function.} Smoothing methods have associated several parameters such as the number of iterations, smoothing, relaxation and so on~\cite{Buja1989}.
\item \textbf{Tuning of the decision rule.} The  $\alpha$-threshold value presented in the decision rule may produce impact in the result performance (accuracy, confusion matrix, etc.).
%
A lower value of $\alpha$ makes a larger interval,  then it is possible to fall in the error of detecting false positive drifts.
On the other, a large value of $\alpha$ means that the method may increase the  false negative error.
That decision rule is commonly used for outlier detection and artifact removal over signals~\cite{BasterrechNCAA2019,BasterrechICAISC2015}.
Note that the mean and standard deviation can be done over a segment of the sequence (for instance last $n$ points), it does not need to be over the full sequence.
In addition, in practice the parameter can be dynamically corrected according to the performance of the matching matrix (in cases when the information about changes on the distribution arrives at certain moment). 
\end{itemize}

\subsection{Algorithm overview}
\begin{figure}[ht]
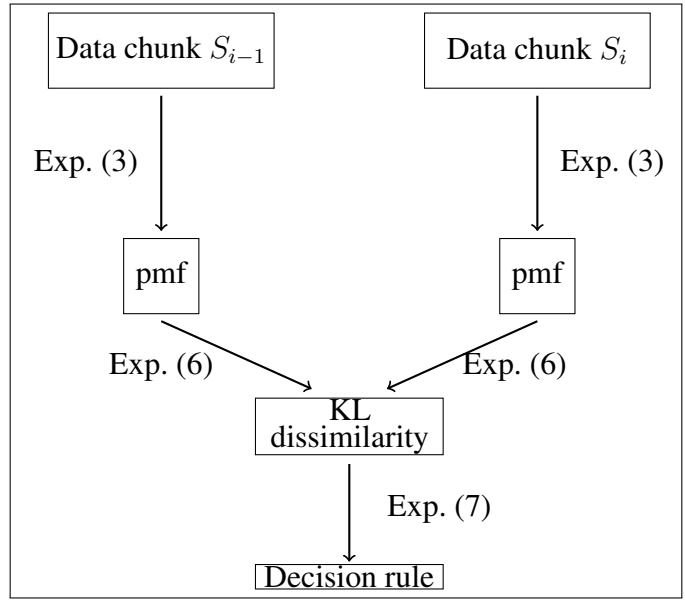

\begin{center}
\fbox{
\scalebox{1}{\tikz{
\tikzstyle{decision} = [diamond, fill=blue!10,  font=\small, text width=4.5em, text badly centered, inner sep=0pt]
\tikzstyle{diam} = [rectangle, draw, font=\small, text width=7em, text badly centered, inner sep=0.5pt]
\draw (1.5,2.5) rectangle (4.5,1.5);
\node (n1) at (3,2) {\large{Data chunk $S_{i-1}$}};
\draw (2.5,-0.5) rectangle (3.5,-1.5);
\node (n2) at (3,-1) {\large{pmf}};
\node (op1) at (2,0.5) {\large{Exp.~(\ref{conditionalProb})}};
\draw (6.5,2.5) rectangle (9.5,1.5);
\node (n2) at (8,2) {\large{Data chunk $S_{i}$}};
\draw (7.5,-0.5) rectangle (8.5,-1.5);
\node (n2) at (8,-1) {\large{pmf}};
\node (op1) at (9,0.5) {\large{Exp.~(\ref{conditionalProb})}};
\draw (3,1.4) edge [->,thick] (3,-0.4);
\draw (8,1.4) edge [->,thick] (8,-0.4);
\node [diam] at  (5.5,-3) {\large{KL dissimilarity}};
\draw (3,-1.6) edge [->,thick] (5,-2.5);
\draw (8,-1.6) edge [->,thick] (6,-2.5);
\node (op2) at (3,-2.2) {\large{Exp.~(\ref{weightedDistanceChunks})}};
\node (op3) at (7.7,-2.2) {\large{Exp.~(\ref{weightedDistanceChunks})}};
\node [diam] at  (5.5,-5) {\large{Decision rule}};
\node (op4) at (6.7,-4.1) {\large{Exp.~(\ref{DecisionRule})}};
\draw (5.5,-3.5) edge [->,thick] (5.5,-4.8);
}}}
\end{center}
\caption{High-level  flowchart of KLD algorithm~\ref{Algo1}.}
\label{Flowchart}
\end{figure}

We present a pseudo-code of the KLD detector in Algorithm~\ref{Algo1}.
The presented version is online, it means given a new chunk performs an evaluation (it checks if there is a change in the distribution).
Although, it is enough few intuitive changes  to transform the online version to an offline version.
Both lines 4 and 6 apply expression~(\ref{conditionalProb}) for estimating the probability mass function.
In this estimation is necessary to define a regular grid of the input space.
This grid can be specifically created for the data on the consecutive two chunks ($S_i$ and $S_{i+1}$) (partitioning a region of $\U$) or it can be made just once for the whole data (partitioning $\U$).
Numerical computation of KL divergence is more stable when a regular grid is created specifically for each pair of consecutive chunks.
In line 10 is applied the smoothing mapping. In our experiments, we didn't evaluate several smoothing methods, we applied moving averages and LOWESS. 
Further research can be done for analysing the impact of the smoothing mapping computation.
Figure~\ref{Flowchart} presents a high-level pipeline of algorithm~\ref{Algo1}. It shows the sequence of the main operations. Each arrow represents a mathematical expression. For simplification, some operations such as the domain partition and smoothness of the dissimilarity metric are not presented in the diagram.

\begin{algorithm}[ht]
\SetAlgoLined
\tcp{\textbf{Main global parameters}: number of bins ($J$), $\alpha$ (threshold in decision rule)}

\tcp{Initial sets}
$D=\emptyset$ (similarity distances) \;
$C=\emptyset$ (critical points - concept drift appearances) \;
\tcp{Initial chunk}
Given chunk $S_0$\;
Apply~(\ref{conditionalProb}) for data in $S_0$ \;
 \While{(it arrives a new chunk ($S_i$))}{
    \tcp{Estimation of pmf}
    Apply~(\ref{conditionalProb}) for data in $S_i$ \;
    \tcp{KL divergence for each bin} 
    Apply~(\ref{distanceBins}) for data in $S_{i-1}$ and $S_i$\; 
    \tcp{Metric aggrregation}
    Apply~(\ref{weightedDistanceChunks}) to compute $D(S_{i-1},S_i)$\; 
$D \leftarrow D\cup D(S_{i-1},S_i)$\;
To apply a smoothing mapping over sequence $D$\;
To compute first derivative (sequence $l$)\;
\tcp{Decision rule}
\If{(expression~(\ref{DecisionRule}) is satisfied)}{
    $C\leftarrow C\cup l_i$\;
 }
 }
Return the set of timestamps of detected drifts (critical points) $C$\;
 \caption{KL-divergence-based concept drift detector (KLD).}
\label{Algo1}
\end{algorithm}

%
%

\section{Experimental study}
\label{ExpStudy}
\subsection{Goal of the experiment}
We designed a set of experiments to answer if we can identify concept drift (critical points) in incrementally drifting data stream only based on topological characteristics of the raw data?

%
\subsection{Experimental setup}
\noindent \textbf{Choice of benchmark data streams. }
As there is a lack of a large diverse collection of real data streams where  concept drift appearances are additionally marked. Therefore, to properly evaluate the proposed methods, a set of simulated data was generated using \emph{Stream-learn} library~\cite{Ksieniewicz:2022}.
Each data stream has~$10000$ chunks with~$250$ instances. We worked over two types of binary datasets with four and six features.
A total of eight simulated data were evaluated. 
Drifts were generated using a sigmoid function that determines how sudden the change of drift concept it is. We used a concepts sigmoid parameter of $99$ (maximum value is 999 - it is the value for simulating a sudden drift). Each dataset has $20$ induced concepts drifts. 
We present in Table~\ref{DataGenerator} the used parameters for generating the data. More details about each of the parameters are specified in~\cite{Ksieniewicz:2022}.
\begin{table}[]
    \centering
    \caption{Applied parameters of the simulated data using  \emph{Stream-learn} library.}
    \label{DataGenerator}
    \begin{tabular}{|l|l|}\hline\hline
    Description & Values \\ \hline
        Seeds & $\{1410, 6543, 2345, 9876, 3946\}$\\
        Number of concept drifts & $20$ \\
        Input features & $\{4,6\}$\\
        Number of classes & $2$\\ 
        Number of chunks & 10000\\
        Chunk size & $250$ \\
        Concept sigmoid spacing & $99$\\
        Class flip & $0.01$ \\\hline\hline
    \end{tabular}

\end{table}

\noindent\textbf{Experimental protocol and implementation.} 
The implementation of the proposed method and the experimental environment are done using the \emph{Python} v3.9 programming language and a few libraries: \emph{NumPy} v1.19.5, \emph{tsmoothie} v1.0.4 and \emph{stream-learn} v0.8.16.  We used the implementation of CART Decision Tree (DT) and Gaussian Na\"ive Bayes Classifier 
form sk-learn v1.0.2, and learning protocol was based on the \emph{Test-Than-Train}~\cite{Krawczyk:2017} evaluation protocol. 

%
%
%
\begin{figure}[ht]
    \centering
    \includegraphics[height=6cm,width=8cm]{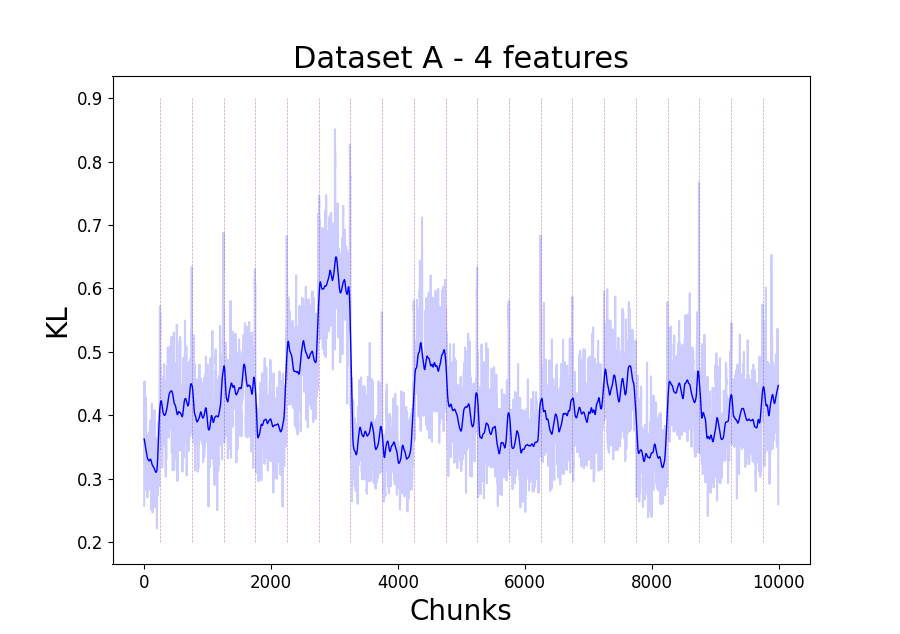}
    \includegraphics[height=6cm,width=8cm]{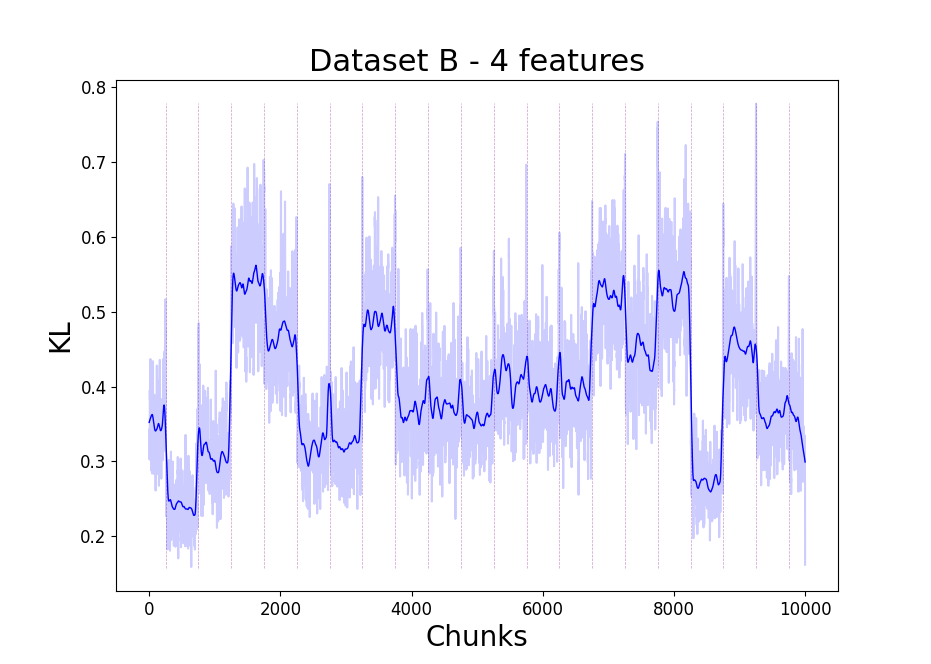}
    \includegraphics[height=6cm,width=8cm]{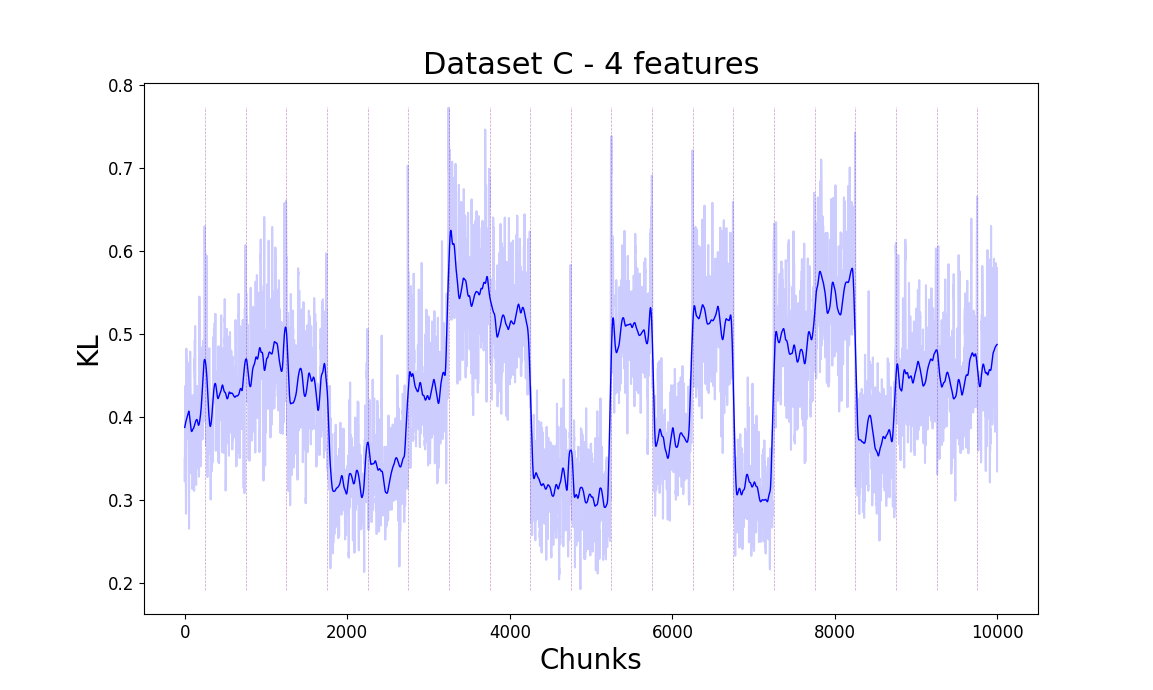}
    \includegraphics[height=6cm,width=8cm]{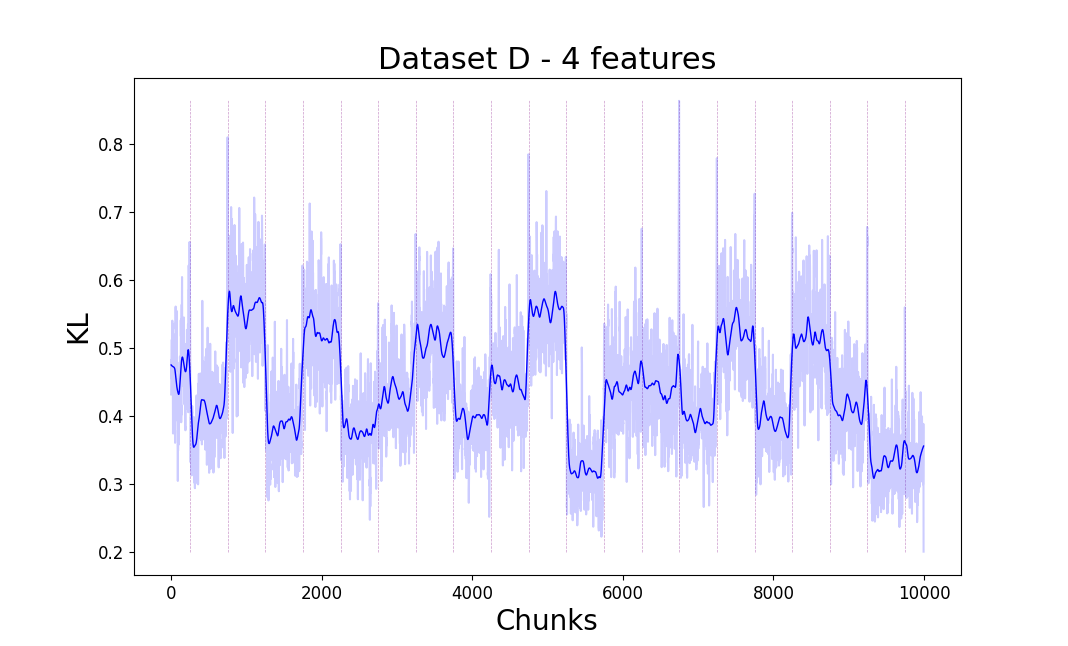}
\caption{KL divergence values for the problems with four features. Dotted vertical lines show where was simulated the data distribution changes.}
    \label{datasetFourfeatures}
\end{figure}    
\begin{figure}[ht]
    \centering
    \includegraphics[height=6cm,width=8cm]{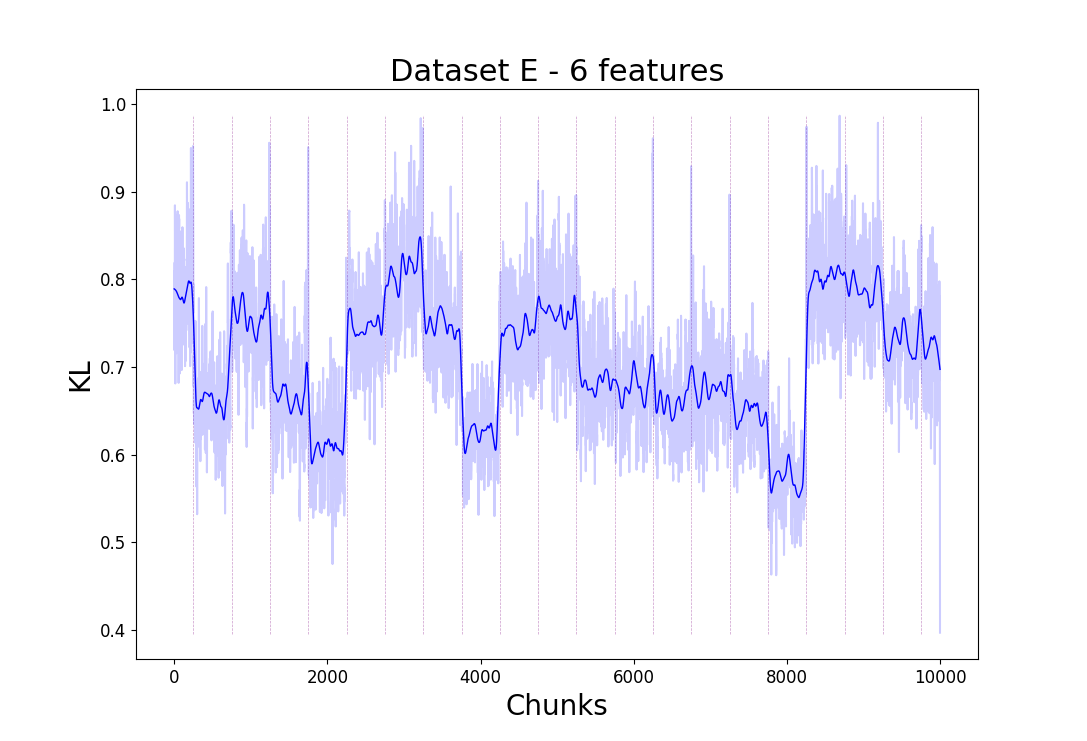}
    \includegraphics[height=6cm,width=8cm]{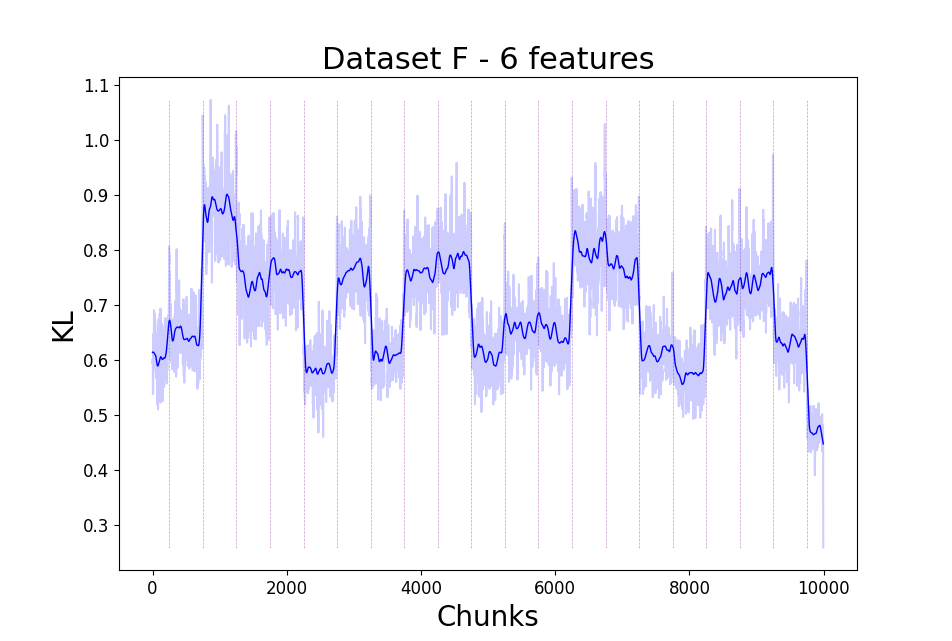}
    \includegraphics[height=6cm,width=8cm]{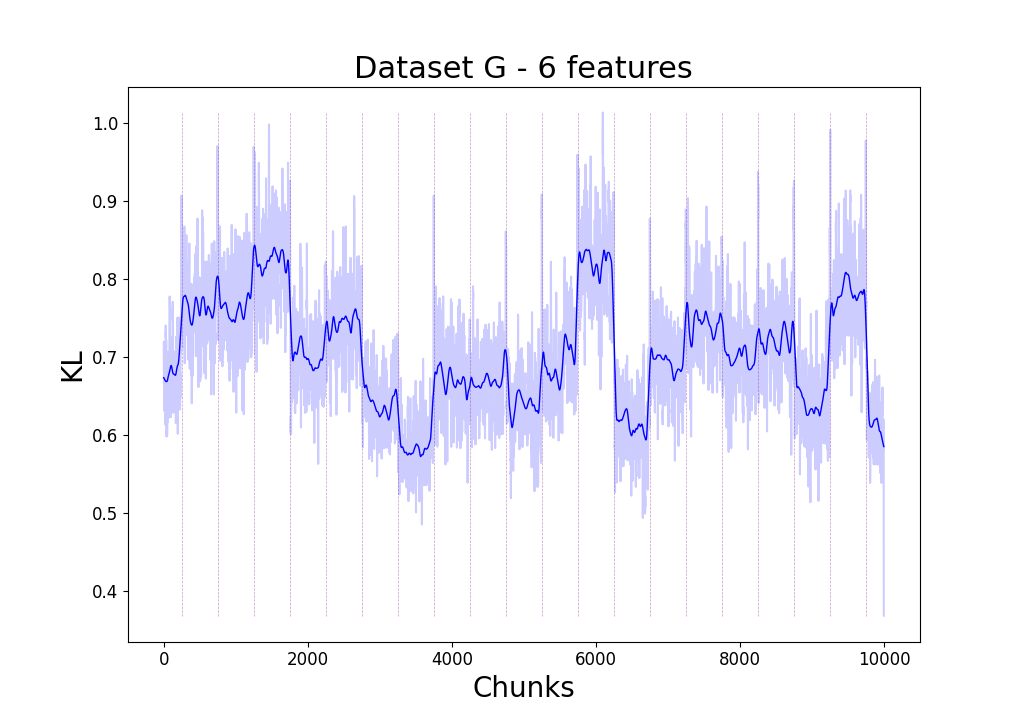}
    \includegraphics[height=6cm,width=8cm]{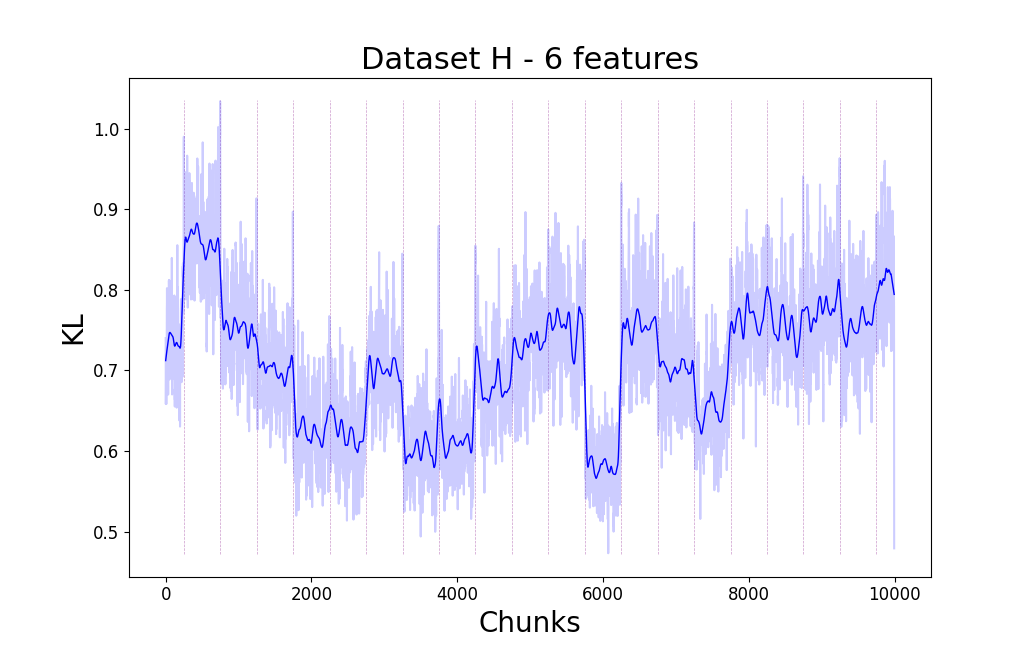}
     \caption{KL divergence values for the problems with six features. Dotted vertical lines show where was simulated the data distribution changes.}
    \label{datasetSixfeatures}
\end{figure}    
\begin{figure}[ht]
    \centering
    \includegraphics[height=6cm,width=8cm]{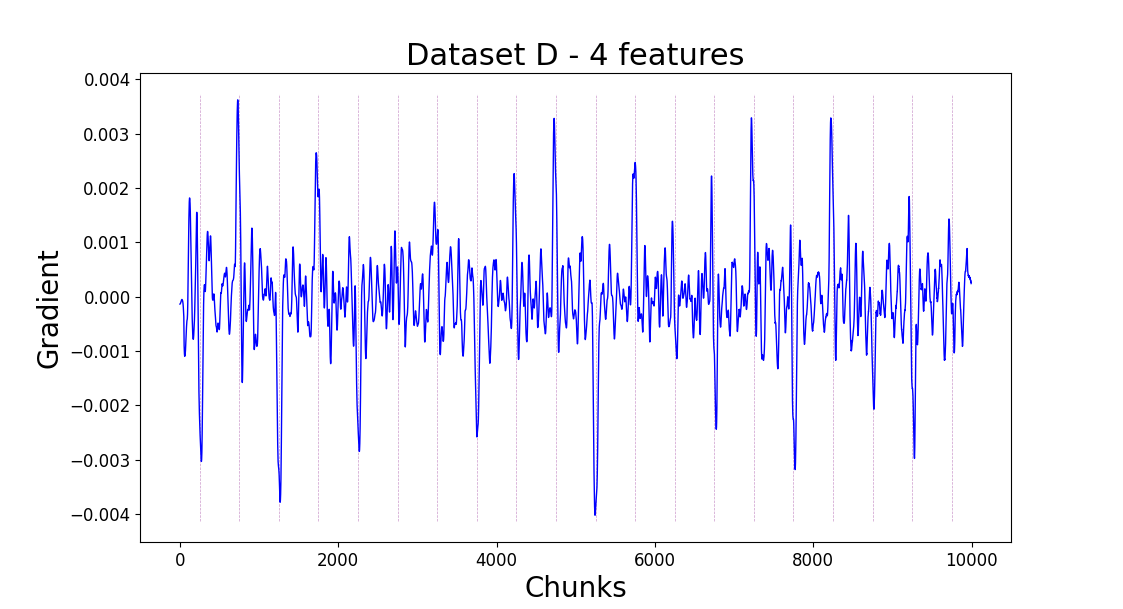} 
    \includegraphics[height=6cm,width=8cm]{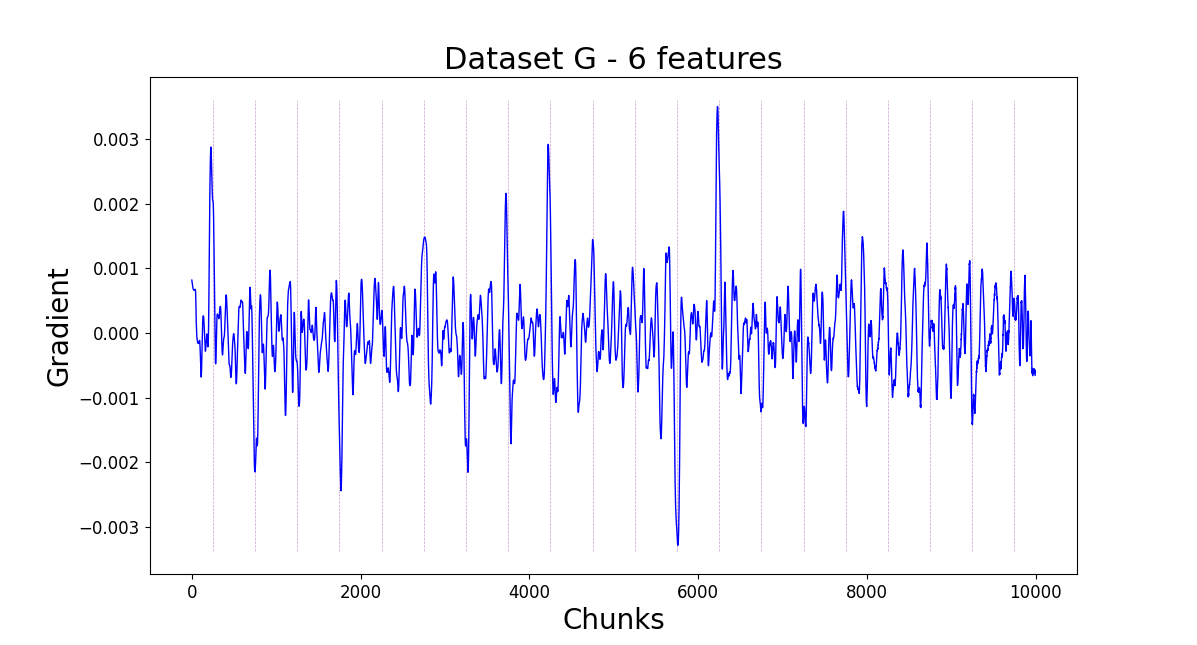}
	\caption{Two examples of the gradient of the smoothed KL divergence curve.}
    \label{Gradient}
\end{figure} 
\begin{figure}[ht]
    \centering
    \includegraphics[height=6cm,width=8cm]{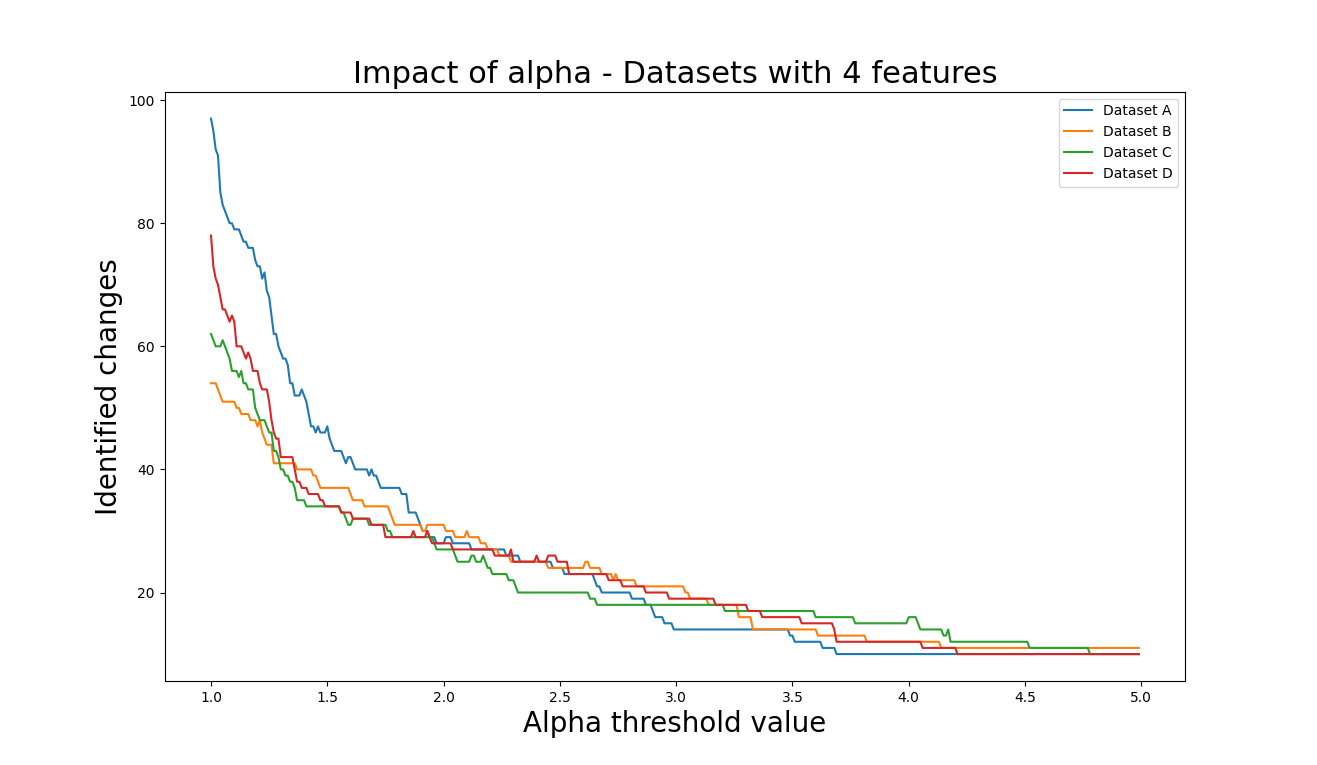} 
    \includegraphics[height=6cm,width=8cm]{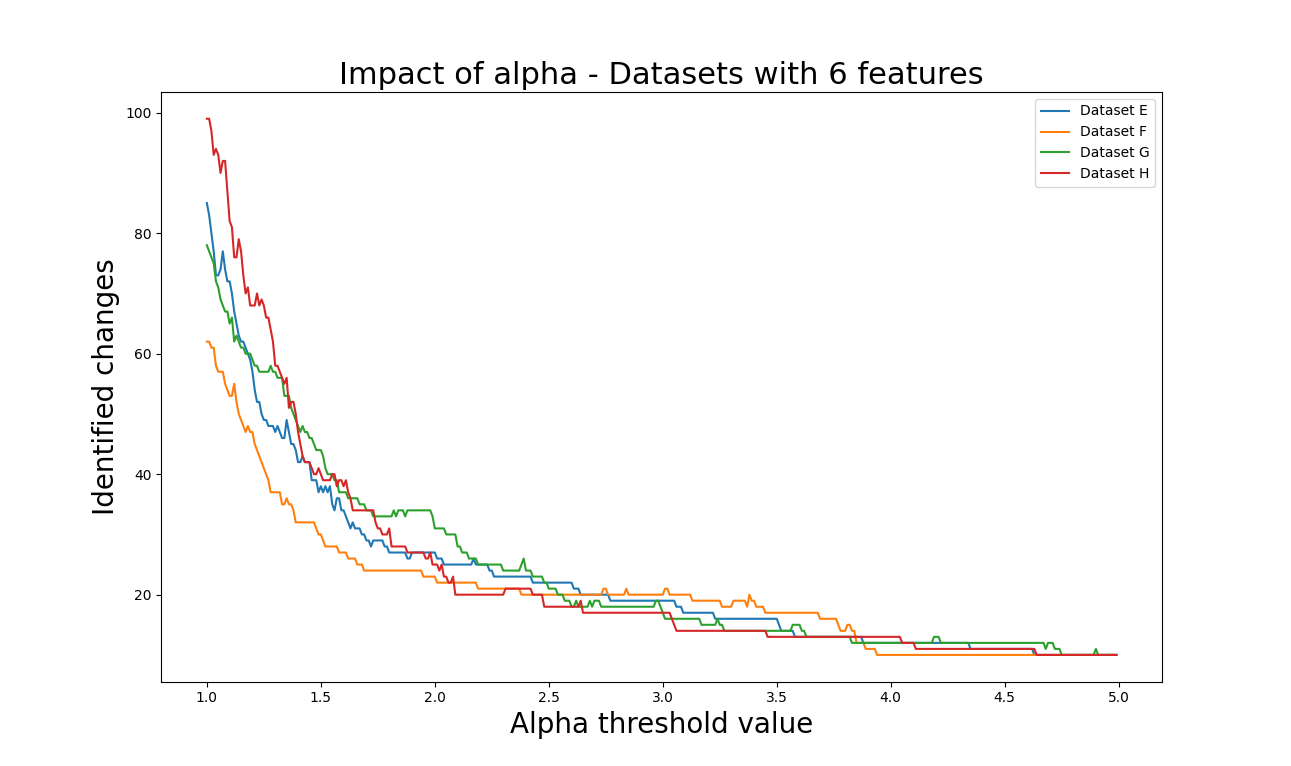}
     \caption{Sensitive analysis of the control parameter $\alpha$. The y-axes show the number of critical points (when the gradient curve surpasses an upper-bound defined by the decision rule).}
    \label{Alpha}
\end{figure}    
\begin{figure}[ht]
    \centering
    \includegraphics[height=6cm,width=8cm]{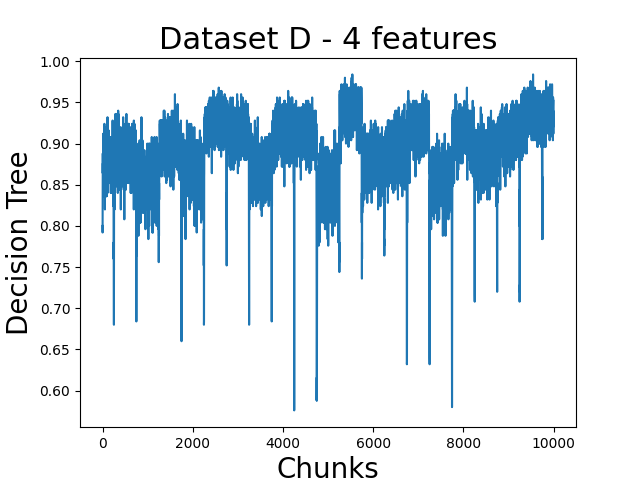}
	\includegraphics[height=6cm,width=8cm]{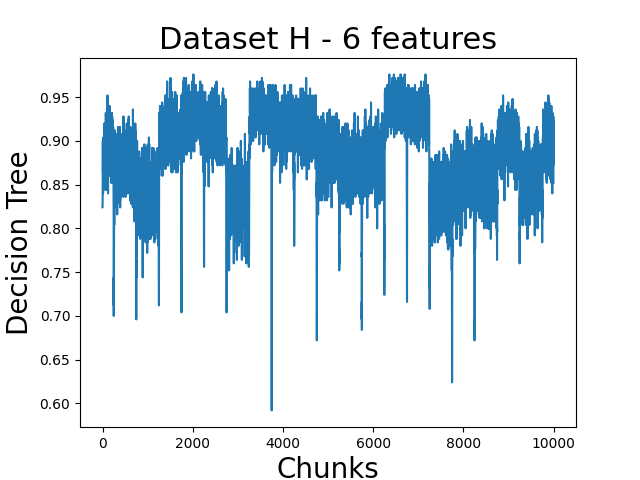}
    \includegraphics[height=6cm,width=8cm]{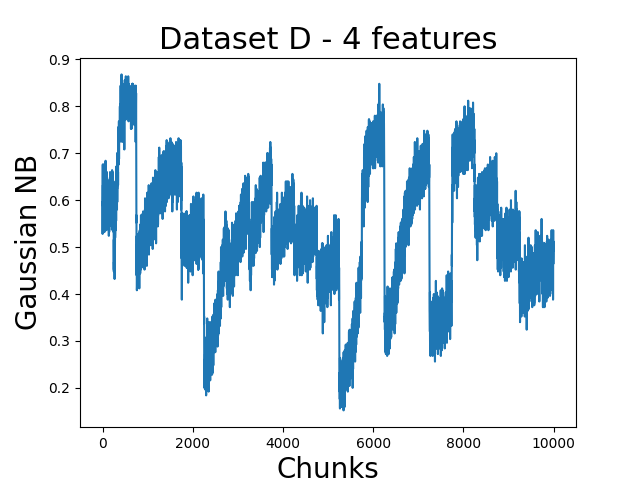}
\includegraphics[height=6cm,width=8cm]{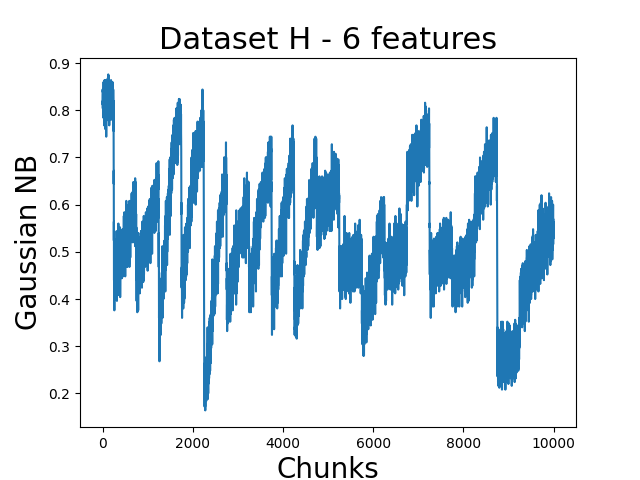}
     \caption{Example of comparison with SOTA techniques. Gradient of the smoothed KL metric curve compared with other tracking techniques (Decision Tree and Gaussian Gaussian Naive Bayes) over datasets with four and six features.}
    \label{VisualComparison}
\end{figure}

\begin{figure}[ht]
    \centering
    \includegraphics[width=11cm]{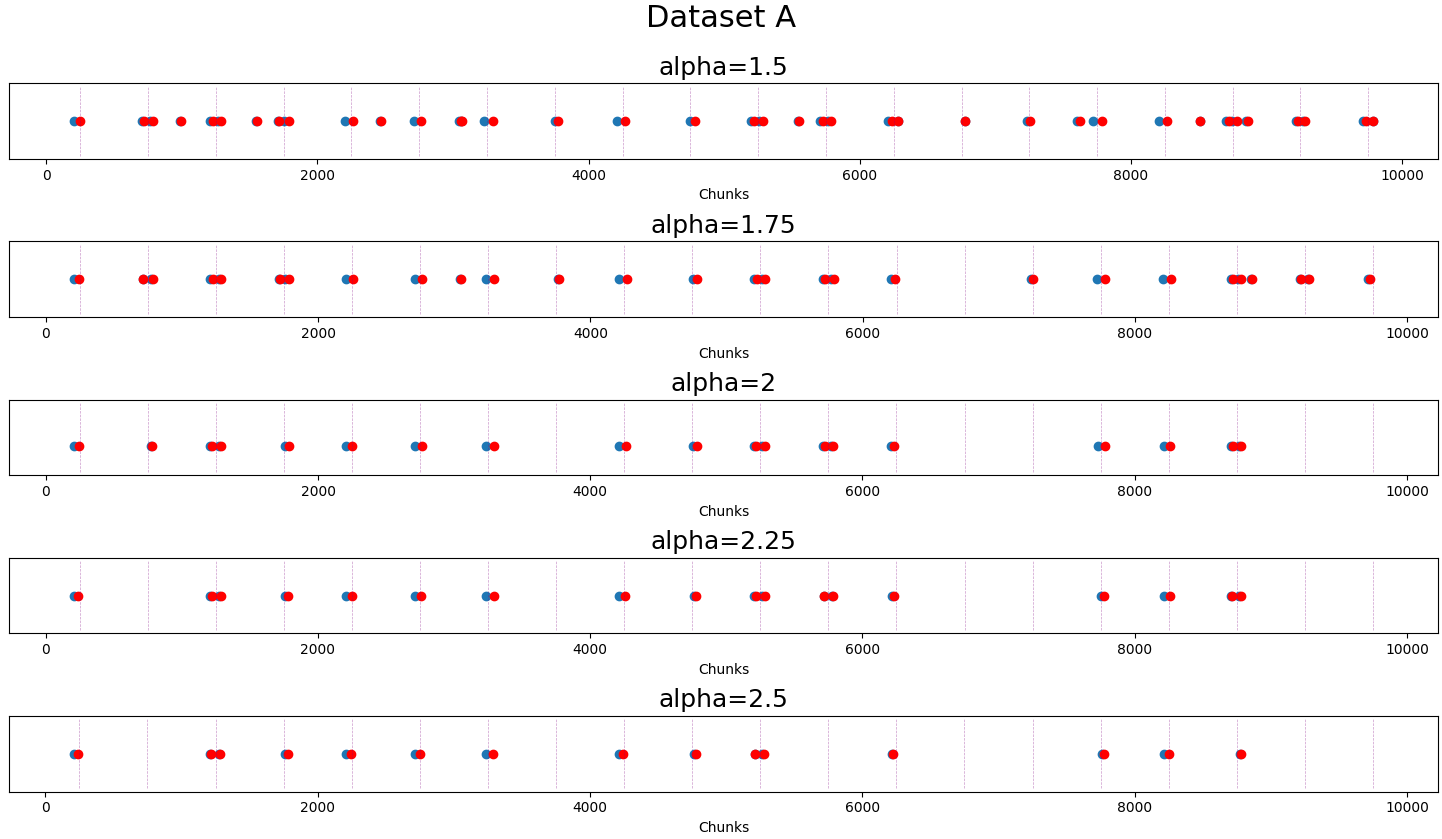} 
    \includegraphics[width=11cm]{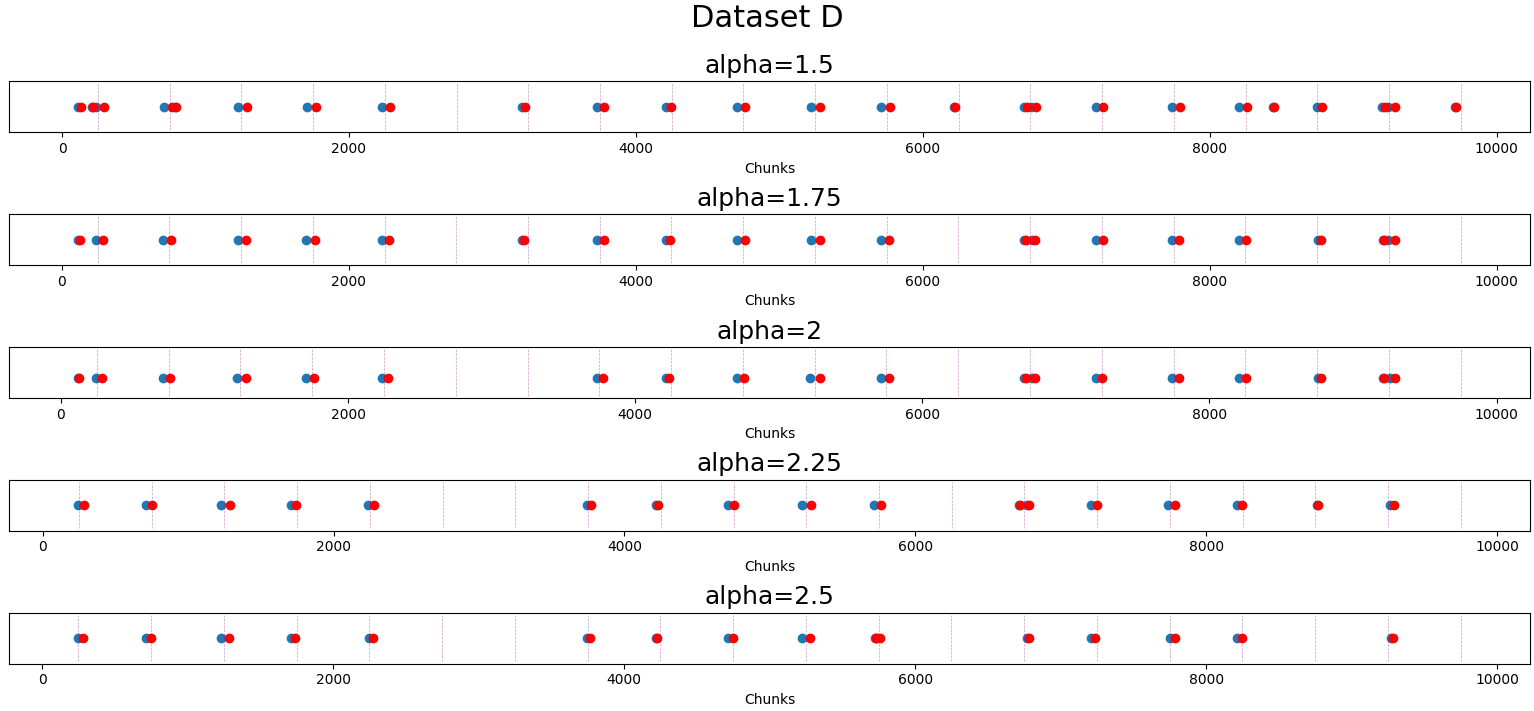}
    \includegraphics[width=11cm]{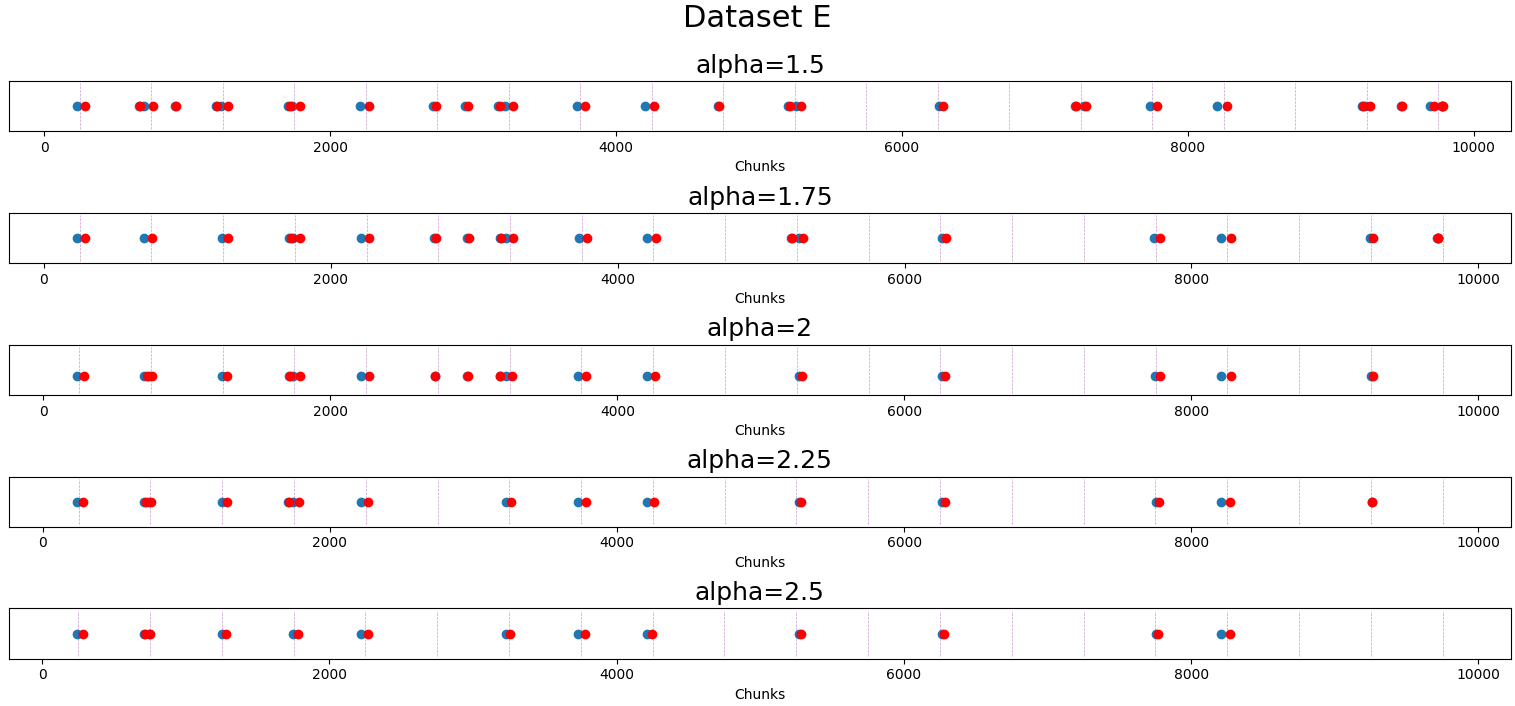}
    \includegraphics[width=11cm]{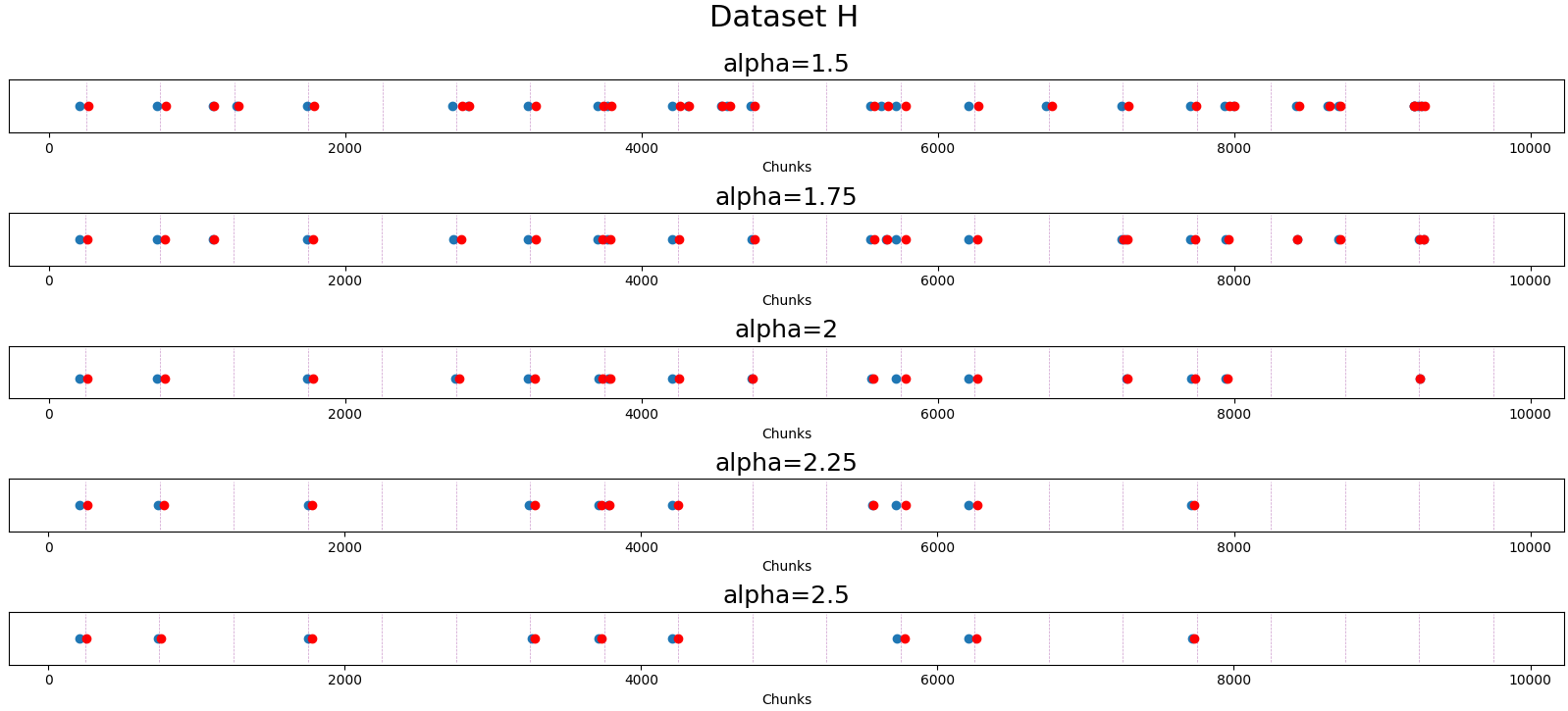}
    \caption{ Sensitive analysis of the control parameter~$\alpha$.
Horizontal axes show the evolution of chunks. Each scatter-plot presents the results obtained with a specific~$\alpha$ value. Vertical lines indicate the timestamps when a concept drift was induced by the simulator. Blue dots represent the timestamps when the gradient curve reaches the upper-bound of the decision rule, and red dots show the moment when the curve goes outside the critical area.}
    \label{AlphaAnalisis}
\end{figure}  

\subsection{Results}

Figure~\ref{datasetFourfeatures} presents four graphics associated to the datasets with four features. 
A similar set of graphics is presented in Figure~\ref{datasetSixfeatures}, in this case the graphics were produced over datasets with six features.
In each graphic, the light blue dots show 
the similarity metric defined in expression~(\ref{distanceChunks}).
The data points are normalized in [0,1].
Over the large cloud of light blue dots, it is shown in dark blue the smoothed curve. This curve is produced using the smoothing mapping (line 10 of Algorithm~\ref{Algo1}).
Furthermore, the graphics have dotted vertical lines that show where is produced the induced drift.
From the figures, it is possible to see the variability of the data, and also to see how the method \textit{properly} reacts to the induced drifts.
Figure~\ref{VisualComparison} presents a visual comparison of the proposed approach with SOTA techniques.
The left group of graphics shows the results for the datasets with four features, and on the right side there are graphics with the results produced over datasets with six features.
It is beyond the scope of this paper to make an exhaustive comparison of the performance of our approach with change-point detection techniques based in classifiers.
KLD only applies similarity analysis over the raw data.
However, we also present an example of visual comparison of our method for monitoring changes and two well-known classifiers (Decision Tree (DT) and Gaussian Naive Bayes (GNB)). 
The two graphics in Figure~\ref{Gradient} are curves with the  first derivative of the smoothed studied metric.
The curves are centered at origin. We can identify extreme values using the decision rule for detecting the critical points.
%
It is worth mentioning that the DT works much better than GNB, it is visible in curves presented in Figure~\ref{VisualComparison}.
In this example, it is also possible to see that KLD can identify change-point locations in places where DT classifier has low accuracy (e.g. in dataset D chunk after 4000, in dataset G chunk before 4000).
%
%
Figure~\ref{Alpha} presents a sensitive analysis of $\alpha$-threshold. The figure has two graphics, one corresponds to the datasets with four features and another one to the datasets with six features.
The number of changes is presented in the  y-axes, we consider a change as the critical point when the gradient curve crosses an upper-bound threshold (defined by the decision rule).
A larger value of $\alpha$ implies that the method increases the number of identified critical point locations.
Note that, the simulated data has $20$ induced drifts. 
An $\alpha$ value lower than $1.2$ produces several false positives, and a large value of $\alpha$ makes that the method is not able to detect too many critical points. The behavior of the impact of $\alpha$ is similar through the eight analyzed problems.
Figure~\ref{AlphaAnalisis} also depicts the behavior of the $\alpha$-threshold.
We mark the segments when the gradient curve reaches the critical interval (interval defined by the decision rule).
The segments start with a blue dot and end with a red dot, i.e., blue dots represent the timestamps
when the gradient curve reaches the critical area, and red dots show the timestamps when the curve escapes from the critical area.
For illustrative purposes, we selected two datasets with four features (A and D) and two datasets with six features (E and H). We show the results of $\alpha\in\{1.5,1.75,2,2.25,2.5\}$.
Each scatterplot has in the x-axes the evolution of chunks.
As in the previous graphics, vertical
lines indicate the timestamps when an incremental concept drift was induced by the simulator. 
According to the graphics seems that an optimal value of $\alpha$ is in the interval $[1.5,1.75]$.
The proposed method has a high ability to detect the change-points locations, and for $\alpha$ values larger than $1.5$ the method almost does not predict false positives.
\subsection{Lessons learned}
Achieved results answered our initial experimental goal.
We have introduced a  fast approach for tracking changes in a data stream, which is based on entropy and similarity analysis over the raw data.
An advantage of the proposal is that the method is universal, in the sense that, it  does not assume any type of specific data distribution.
The technique has a parameter ($\alpha$) that is easy to adjust, and directly impacts the matching matrix.
We empirically found a short interval where the $\alpha$ value can be tuned.
Empirical results show that, independently of the input domain, it is suitable to apply the method with $\alpha$ value in $[1.5,2]$.
\section{Conclusions and future work}
We presented a fast and effective method for tracking changes in probability distributions that may occur during continuous learning. It should be noted that the presented evaluation results of the proposed method are preliminary. However, they confirm its usefulness, which encourages us to continue the work in the future.
The most important directions for further work include: 
(i) Evaluate the effectiveness over other data domains. Current approach has been evaluated only on binary output variables.
(ii) Applying it to real-world continual learning tasks with different type of drifts.
(iii)  Analysis of its suitability for processing visual information, i.e., image streams with so-called syntactic drift.
(iv)  Consider other similarity measures such as Jensen-Shannon entropy or Wasserstein distance.
\clearpage
\bibliographystyle{plain}
\bibliography{References}






\end{document}